\documentclass[10pt,journal,compsoc]{IEEEtran}

\usepackage{cite}
\usepackage{multirow}
\usepackage{subfigure}
\usepackage{amssymb}
\usepackage{xspace}
\usepackage{epsf}
\usepackage{setspace}
\usepackage{caption}
\usepackage{tikz}
\usepackage{booktabs}
\usepackage[vlined,linesnumbered,ruled]{algorithm2e}
\usepackage{balance}
\usepackage{url,epsfig,array}
\usepackage{amsmath}
\usepackage{epstopdf}
\usepackage{cases}
\usepackage{bm}
\usepackage{amsthm}
\usepackage{enumitem}
\theoremstyle{definition}
\def\ie{\textit{i.e.}\xspace}

\def\eg{\textit{e.g.}\xspace}

\def\method{ReLoRA\xspace}

\usepackage{supertabular}
\usepackage{pifont}

\begin{document}

\title{\textit{\method}: Knowledge-Reusing Adaptation for Fast Rollout of Evolving LLM Services}

\author{
    \IEEEauthorblockN{
        Yang Xu,~\IEEEmembership{Member,~IEEE,}~
        Zihuai Xu,~
        {*}Hongli Xu,~\IEEEmembership{Member,~IEEE,}~
        Yunming Liao,~
        Zhiwei Yao,~
        Xitong Fu\\
    }
    \IEEEcompsocitemizethanks{
        \IEEEcompsocthanksitem This article is supported by the National Science Foundation of China (NSFC) under Grants XXXXXXXX, XXXXXXX, and XXXXXXX.
        \IEEEcompsocthanksitem Y. Xu, Z. Xu, H. Xu, Y. Liao, Z. Yao, and X. Fu are with the School of Computer Science and Technology, University of Science and Technology of China, Hefei, Anhui, China, 230027, and also with Suzhou Institute for Advanced Research, University of Science and Technology of China, Suzhou, Jiangsu, China, 215123. \protect E-mails: xuyangcs@ustc.edu.cn; zihuaixu@mail.ustc.edu.cn; xuhongli@ustc.edu.cn; ymliao98@mail.ustc.edu.cn; zhiweiyao@mail.ustc.edu.cn; xt1310632375@mail.ustc.edu.cn.
        \IEEEcompsocthanksitem H. Xu is the corresponding author.
    }
}

\markboth{IEEE Transactions on Services Computing, Vol., No., MAY. 2026}%
{Shell \MakeLowercase{\textit{et al.}}: Bare Advanced Demo of IEEEtran.cls for Journals}

\IEEEtitleabstractindextext{%
\begin{abstract}
Large Language Models (LLMs) are increasingly deployed as continuously evolving services, where frequent base-model updates may invalidate previously deployed task-specific Low-Rank Adaptation (LoRA) adapters. For service providers managing numerous downstream model services, retraining each LoRA adapter from scratch for every updated base model is computationally prohibitive and delays service rollout. Meanwhile, the simpler alternative, \ie, naively applying the original LoRA adapter to the updated base model, often leads to degraded service quality due to adapter-backbone incompatibility. To address this problem, we propose \method, a knowledge-reusing re-adaptation framework that efficiently restores service-ready LoRA adapters for evolving LLM services while preserving or improving task performance. Specifically, \method\ comprises two key optimization steps: 1) \textbf{Adaptive LoRA initialization} leverages Bayesian optimization to construct a compatibility-aware starting point by fusing information from both the previously deployed task adapter and the base model's evolution; 2) \textbf{Fine-tuning with scheduled regularization} first rapidly steers the adapter to a high-quality region via strong regularization, followed by relaxed regularization for task-specific refinement. This design enables rapid service-quality recovery with reduced re-adaptation overhead. Extensive experiments demonstrate that \method\ reduces time-to-readiness by up to 8.9$\times$ and improves accuracy by up to 4.6\% compared to baselines.

\end{abstract}

}

\maketitle

\IEEEdisplaynontitleabstractindextext

\IEEEpeerreviewmaketitle

\ifCLASSOPTIONcompsoc
\IEEEraisesectionheading{\section{Introduction}\label{sec:intro}}
\else
\section{Introduction}\label{sec:intro}
\fi
Large Language Models (LLMs), such as ChatGPT~\cite{wu2023brief} and LLaMA~\cite{touvron2023llama}, have become a fundamental component of modern AI services~\cite{cao2023comprehensive}. 
Through extensive pretraining on diverse and massive corpora, these models exhibit remarkable generalization capabilities across a wide range of tasks. 
Increasingly, LLMs are not deployed as static models, but as continuously evolving services whose base models are periodically updated to incorporate new knowledge, improve alignment, and enhance general capabilities. 
In such service-oriented ecosystems, a provider-side base-model update is not merely a model replacement; it also triggers a lifecycle maintenance problem for numerous downstream model services built upon the previous model version.

To support specialized downstream applications, LLM services are commonly customized through fine-tuning. 
However, conventional full-parameter fine-tuning requires substantial computation and memory overhead. 
For instance, fine-tuning a LLaMA model with 13 billion parameters requires approximately 100GB of memory. 
To reduce this cost, Parameter-Efficient Fine-Tuning (PEFT) methods have been developed, among which Low-Rank Adaptation (LoRA)~\cite{hu2021lora} has become a predominant solution. 
LoRA freezes the base pretrained model weights and represents task-specific updates through low-rank matrix decomposition, significantly reducing resource demands while maintaining competitive performance. 
As a result, many downstream LLM services are deployed as a shared base model equipped with task-specific LoRA adapters.

However, the continuous evolution of LLM backbones, such as Gemini~\cite{team2023gemini} and LLaMA~\cite{touvron2023llama}, poses a critical challenge to such LoRA-based service deployment. 
When the base model evolves from an old version to a new one, previously deployed LoRA adapters may become obsolete or suboptimal because they were optimized for the old backbone. 
Consequently, service providers who have invested in fine-tuning task-specific adapters may have to update these adapters before the corresponding services can be safely rolled out on the evolved base model. 
For service providers managing thousands of LoRA-based task services in production~\cite{aws_sagemaker,togetherai}, even a moderate per-adapter retraining cost can accumulate into substantial GPU-hours for each base-model update cycle. 
For instance, if fine-tuning a single LoRA instance on a moderately sized dataset takes approximately 4 hours on an NVIDIA A100 GPU~\cite{dettmers2023qlora}, updating 2,000 LoRA instances would require roughly 8,000 GPU hours, resulting in a significant operational burden~\cite{gao2024cost,aws_pricing}. 
This cost directly delays the rollout of updated LLM services and increases the maintenance overhead.

To mitigate this problem, existing approaches have largely followed two distinct paths. 
The most straightforward implementation, called PortLLM~\cite{khan2024portllm}, directly applies the original LoRA weights to the newly evolved base model. 
From a service-management perspective, this approach is appealing because it requires almost no additional computation and can immediately reuse existing task-specific knowledge. 
However, as we demonstrate in Section~\ref{subsec: motivation}, such naive adapter transfer fails to account for the parameter shifts introduced by base-model evolution. 
This causes a mismatch between the original task adaptation and the evolved backbone, leading to degraded service quality. 
Another line of research explores generative methods to synthesize LoRA weights for evolving LLMs. 
For instance, ORAL~\cite{khan2025oral} employs conditional recurrent diffusion models to generate LoRA weights from scratch based on task specifications and model architecture. 
Although such generative approaches provide flexibility in creating adapters for new tasks, they typically require substantial upfront investment in training the generative model itself, and their effectiveness highly depends on the quality of the learned generator and conditioning information.

Given that direct adapter transfer may lead to unacceptable service-quality degradation, some form of re-adaptation is necessary. 
Meanwhile, to enable fast service rollout, re-adaptation should avoid retraining each adapter from scratch. 
Our key insight is that two forms of existing knowledge can be jointly reused to accelerate this process: the task-specific knowledge encoded in the previously deployed LoRA adapter, and the evolution knowledge reflected by the parameter shift between the old and updated base models. 
By properly integrating these two sources of knowledge, the updated adapter can start from a more compatible parameter region, thereby reducing the time required to restore service-ready performance.

Building on this insight, we propose \method, a knowledge-reusing re-adaptation framework for fast rollout of evolving LLM services. 
\method\ orchestrates the adapter update process through two key steps, \ie, \textbf{adaptive LoRA initialization} and \textbf{fine-tuning with scheduled regularization}. 
First, adaptive LoRA initialization constructs a compatibility-aware starting point by fusing information from the original task adapter with the delta of the base model's evolution, \ie, the difference between the evolved and original base models. 
This fusion is achieved by searching for an optimal weighted combination via Bayesian optimization, which provides a superior initialization for subsequent re-adaptation. 
Second, fine-tuning with scheduled regularization efficiently steers the adapter toward a new task-specific optimum while reducing re-adaptation overhead. 
This step begins with a \textit{Guided Rapid Adaptation} stage, where strong L2 regularization anchors the adapter to the initialized state and rapidly moves it into a high-quality service-ready region. 
It then proceeds to a \textit{Refinement and Exploration} stage, where the regularization is relaxed to allow task-loss-driven refinement for improved final performance. 
Overall, the two-step design enables fast service-quality recovery while preserving downstream task performance.

Our main contributions can be summarized as follows.

\begin{itemize}
\item We formulate adapter re-adaptation for evolving LLMs as an adapter backward-compatibility and service rollout problem, where the objective is to rapidly restore service-ready task adapters after base-model evolution.

\item We propose \method, a knowledge-reusing framework that reduces the maintenance overhead of LoRA-based LLM services by leveraging both the previously deployed task adapter and the base model's evolution.

\item We introduce an adaptive LoRA initialization strategy that employs Bayesian optimization to fuse knowledge from the existing LoRA adapter and the base-model evolution delta, thereby constructing a compatibility-aware starting point for fast re-adaptation.

\item We design a fine-tuning strategy with scheduled regularization, which first anchors the optimization process to the fused initialization for rapid service-quality recovery and then relaxes the constraint for refinement.

\item We conduct extensive experiments across six downstream service tasks, three model families, and three update sources. 
The results show that \method\ reduces time-to-readiness by up to 8.9$\times$ and improves task accuracy by up to 4.6 percentage points over baselines.
\end{itemize}

The rest of the paper is organized as follows. 
Section~\ref{sec:prelim} introduces the background of LoRA and the motivation for this work. 
Section~\ref{sec:algorithm} details the proposed \method\ framework. 
Section~\ref{sec:evaluation} presents the experimental results and analysis. 
Section~\ref{sec:relatedwork} reviews the related works. 
Section~\ref{sec:limitation} discusses limitations of our work, and Section~\ref{sec:concluesion} concludes the paper.

\begin{figure}[!t]
\centering
    \includegraphics[width=1.0\columnwidth]{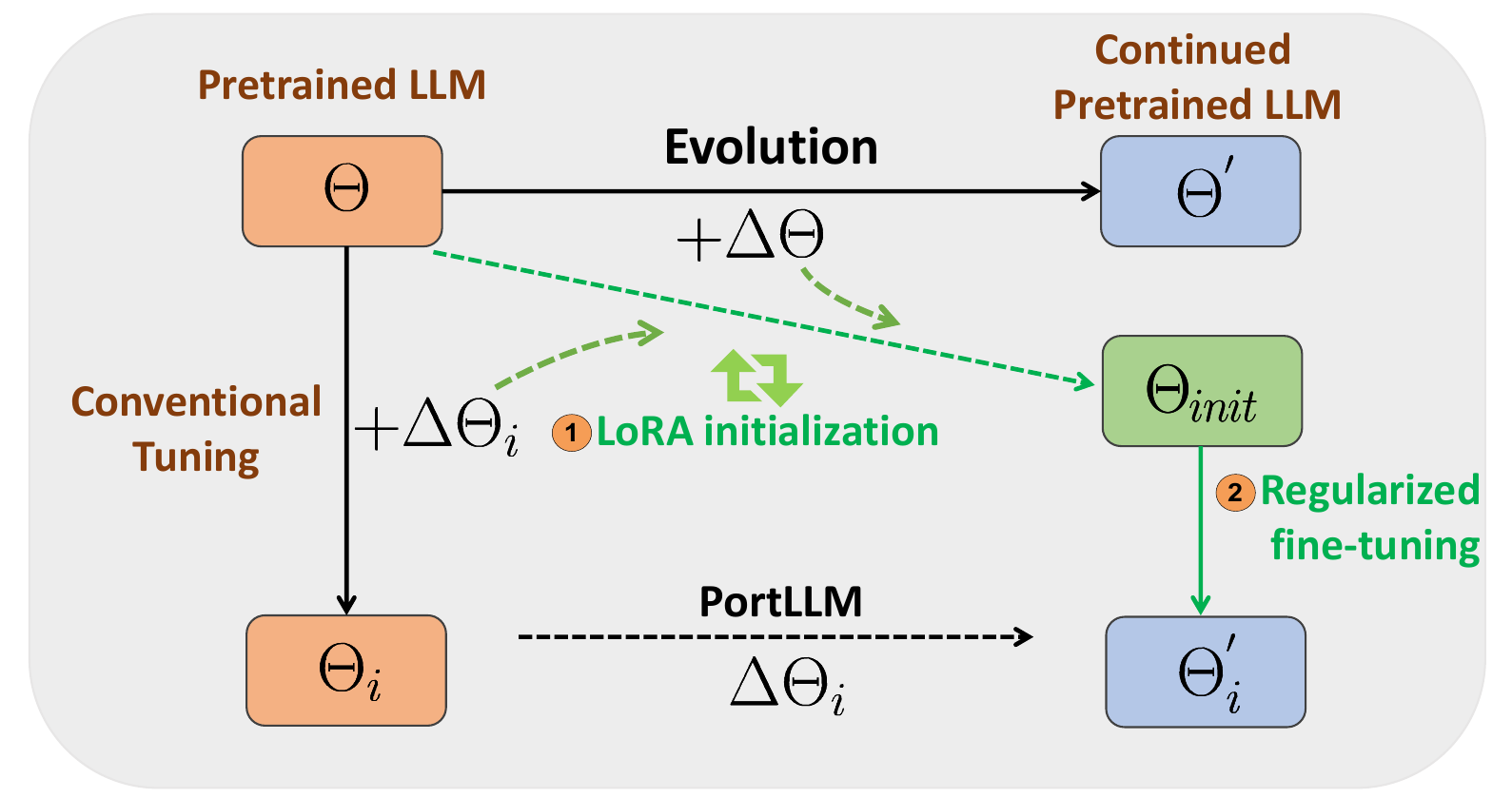}
\caption{Workflow of \method for fast rollout of evolving LLM services.} 
\label{fig: overview}
\end{figure}

\vspace{-3mm}
\section{Preliminaries and Motivations}\label{sec:prelim}
\subsection{Low-Rank Adaptation (LoRA)}
\label{subsec: LoRA}

Parameter-Efficient Fine-Tuning (PEFT) methods substantially reduce the number of trainable parameters required to adapt LLMs to downstream tasks. 
Among them, Low-Rank Adaptation (LoRA) has become a widely adopted technique for customizing LLM services, as it keeps the pretrained backbone frozen and updates only a small set of additional adapter parameters. 
This design allows downstream model services to share the same base model while maintaining separate lightweight task-specific adapters.

Specifically, for a weight matrix $\Theta_0^h \in \mathbb{R}^{d \times k}$ at the $h$-th layer of an LLM, LoRA represents the incremental update for the $i$-th downstream task as a low-rank decomposition:
\begin{equation}
    \Delta\Theta_i^h = B_i^h A_i^h ,
\end{equation}
where $B_i^h \in \mathbb{R}^{d \times r}$, $A_i^h \in \mathbb{R}^{r \times k}$, and the rank satisfies $r \ll \min\{d,k\}$. 
Given an input representation $x$, the layer output is computed as
\begin{equation}
    \Theta_0^h x + B_i^h A_i^h x .
\end{equation}
During fine-tuning, the original weight matrix $\Theta_0^h$ remains fixed, while only $A_i^h$ and $B_i^h$ are trainable. 
For simplicity, we use $\Delta\Theta_i$ to denote the collection of LoRA updates across all adapted layers for the $i$-th downstream service task.

\subsection{LLM Service Evolution and Adapter Backward Compatibility}
\label{subsec:problem_formulation}

We next formalize the adapter backward-compatibility problem for evolving LLM services. 
Let $\Theta$ denote the old service backbone and $\Theta' = \Theta + \Delta\Theta$ denote the updated backbone after base-model evolution, where $\Delta\Theta$ represents the parameter delta introduced by the update. 
For the $i$-th downstream service task, let $\Delta\Theta_i$ denote the previously deployed LoRA adapter on $\Theta$, and let $\Delta\Theta'_i$ denote the updated adapter to be constructed for $\Theta'$. 
The service quality of the updated task service is denoted by $Q_i(\Theta', \Delta\Theta'_i)$.

The objective is to minimize the \emph{time-to-readiness} $\tau_i$, \ie, the wall-clock time from the completion of the base-model update to the point where the updated adapter reaches the required service-quality threshold $q_i$:
\begin{equation}
    \min_{\Delta\Theta'_i} \tau_i 
    \quad 
    \text{s.t.} 
    \quad 
    Q_i(\Theta', \Delta\Theta'_i) \ge q_i .
\end{equation}
This formulation differs from conventional fine-tuning. 
Instead of only maximizing final task accuracy, the key objective is to rapidly restore a service-ready adapter after base-model evolution while preserving service quality.

\subsection{Challenges in Adapting LoRA for Evolving LLM Services: The Pitfall of Naive Transfer}
\label{subsec: pitfall}

Although LoRA significantly lowers the cost of downstream adaptation, updating a large number of LoRA-based services after every base-model evolution remains expensive. 
This issue is especially important in LLM service ecosystems, where multiple task-specific adapters are commonly deployed on top of a shared backbone. 
When the backbone evolves, retraining every adapter from scratch can reliably recover task performance, but it also delays the rollout of downstream services. 
A more attractive strategy is to directly reuse the previously deployed adapter on the updated backbone, thereby avoiding re-training.

PortLLM~\cite{khan2024portllm} follows this direction by treating the task-specific adapter $\Delta\Theta_i$ learned on the old backbone as a transferable ``patch''. 
As illustrated in Figure~\ref{fig: overview}, when the base model evolves from $\Theta$ to $\Theta'$, where
\begin{equation}
    \Theta' = \Theta + \Delta\Theta ,
\end{equation}
PortLLM directly applies the original adapter $\Delta\Theta_i$ to the updated backbone. 
The resulting adapted model can be written as
\begin{equation}
    \Theta'_i = \Theta' + \Delta\Theta_i 
    = \Theta + \Delta\Theta + \Delta\Theta_i .
\end{equation}
From a service-management perspective, this strategy enables an almost immediate rollout path after model evolution because it requires no adapter re-training. 
However, this convenience comes with a major risk: the old adapter was optimized for the previous backbone and may no longer be compatible with the evolved one.

The limitation of naive transfer stems from its neglect of the interaction between the base-model evolution $\Delta\Theta$ and the task-specific adaptation $\Delta\Theta_i$. 
These two components may be complementary or conflicting. 
If the base-model update changes internal representations that are important for the downstream task, simply attaching the old adapter to the updated backbone may move the model to a suboptimal parameter region, leading to degraded service quality.

\begin{figure*}[t]
\centering
\subfigure[]{
\centering
\includegraphics[width=0.225\linewidth]{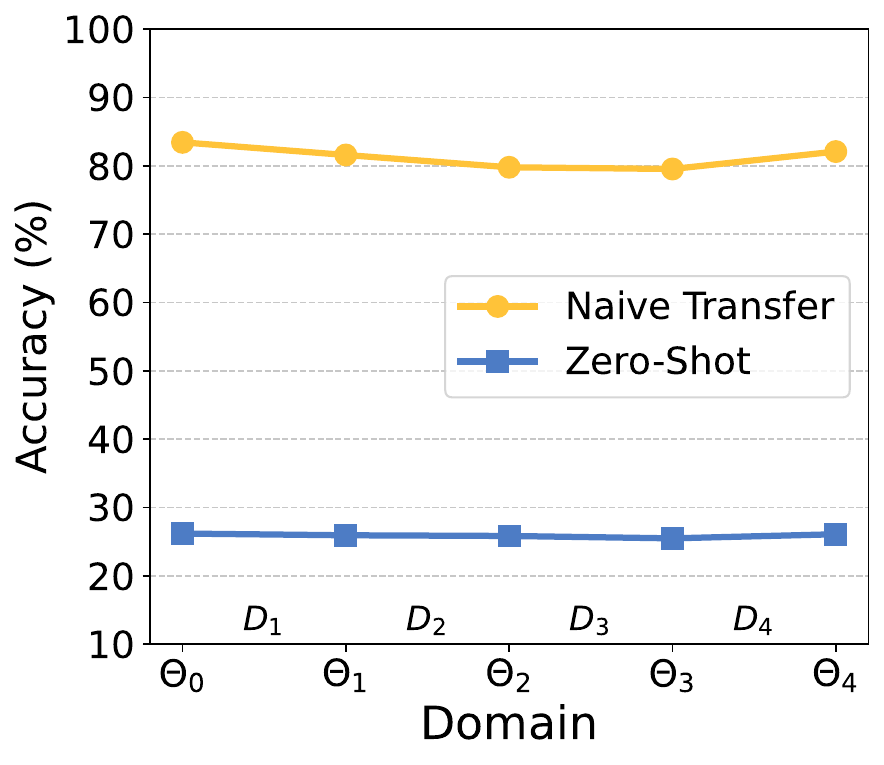}
\label{fig:domain}}
\subfigure[]{
\centering
\includegraphics[width=0.22\linewidth]{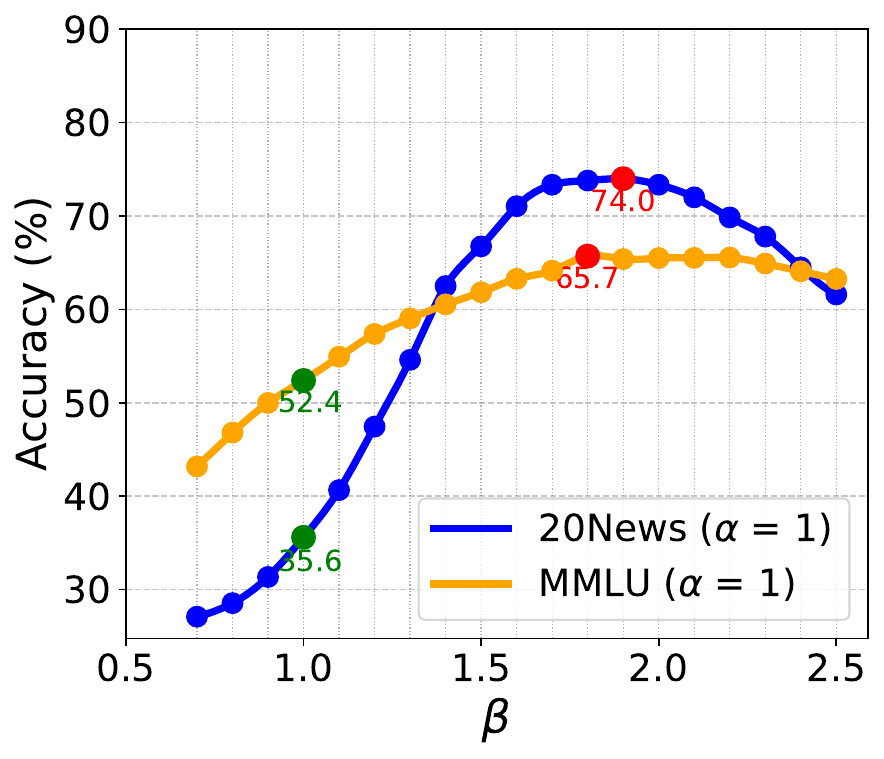}
\label{fig:beta}}
\subfigure[]{
\centering
\includegraphics[width=0.22\linewidth]{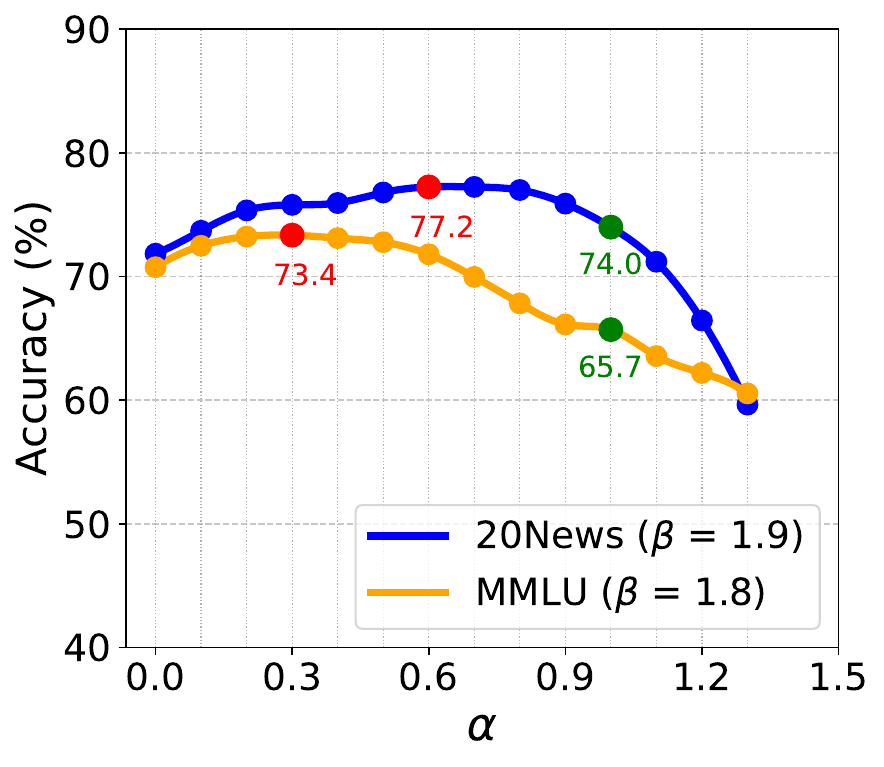}
\label{fig:alpha}}
\subfigure[]{
\centering
\includegraphics[width=0.22\linewidth]{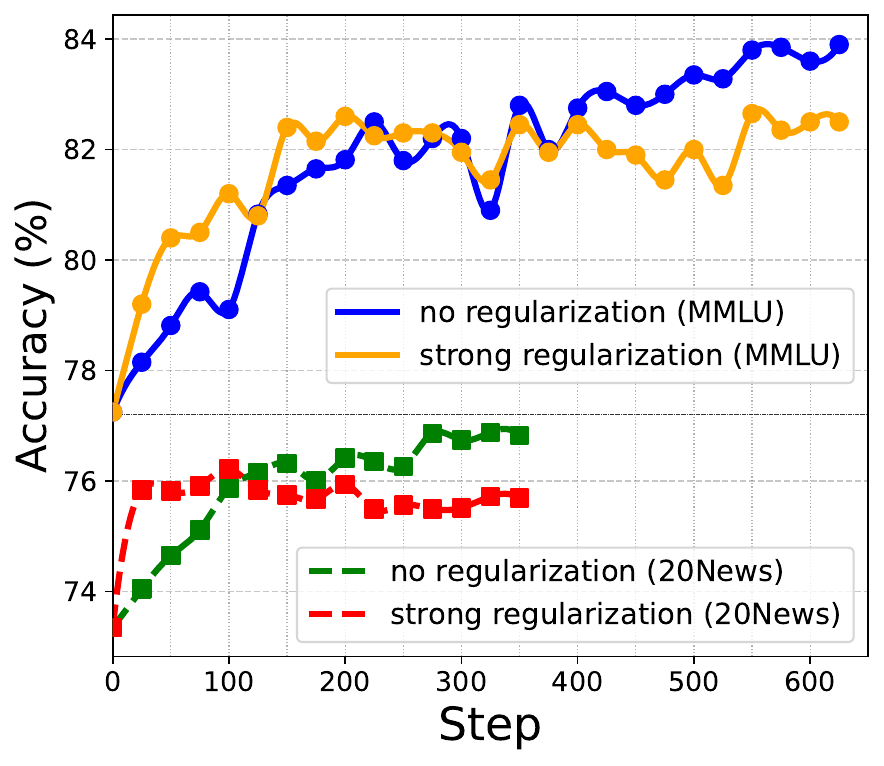}
\label{fig:regularization}}
\caption{The results of preliminary experiments. 
(a) Performance variation when applying the previously fine-tuned LoRA adapter $\Delta\Theta_i$ to $\{\Theta_1,\cdots,\Theta_4\}$; 
(b) Training performance with different $\beta$ values when $\alpha=1$; 
(c) Training performance with different $\alpha$ values; 
(d) Training performance under strong regularization/no regularization on MMLU and 20News.}
\label{fig:prelim}
\end{figure*}

To evaluate the impact of model evolution on naive transfer, we conduct preliminary experiments using LLaMA3.1-8B~\cite{grattafiori2024llama} as the initial base model $\Theta_0$. 
We simulate model evolution through continual pretraining~\cite{qin2023recyclable,qin2022elle} on four distinct domains sequentially, including biomedical papers ($\mathbf{D}_1$)~\cite{lo2019s2orc}, Amazon reviews ($\mathbf{D}_2$)~\cite{he2016ups}, computer science papers ($\mathbf{D}_3$)~\cite{lo2019s2orc}, and news articles ($\mathbf{D}_4$)~\cite{zellers2019defending}. 
Each domain is used for 6k pretraining steps with a batch size of 8. 
Let $\Theta_u$ denote the evolved model after continual pretraining on $\mathbf{D}_u$. 
For downstream evaluation, we use the MMLU benchmark~\cite{hendrycks2020measuring}. 
We directly apply a LoRA adapter $\Delta\Theta_i$, originally fine-tuned on $\Theta_0$, to each evolved model $\Theta_u$ and measure the resulting performance.

As shown in Figure~\ref{fig:domain}, we obtain three observations. 
First, each evolved model equipped with the old adapter performs worse than the original adapted model, indicating that $\Delta\Theta_i$ becomes less compatible as the backbone evolves. 
Second, despite this degradation, reusing the adapter still consistently outperforms zero-shot inference, suggesting that the previously deployed adapter retains useful task-specific knowledge. 
Third, among all evolved models, $\Theta_4+\Delta\Theta_i$ achieves the best performance, although $\Theta_4$ has undergone the longest evolution path. 
This may be because the news domain $\mathbf{D}_4$ is more aligned with the original pretraining distribution of $\Theta_0$~\cite{gururangan2020don}. 
These results show that the effectiveness of a transferred adapter is sensitive to the nature of the base-model evolution. 
Therefore, simply reusing $\Delta\Theta_i$ is insufficient for reliable service rollout.

\subsection{Motivation}
\label{subsec: motivation}

The above observations suggest that the previously deployed adapter remains valuable, but it should not be reused blindly after base-model evolution. 
Naive transfer implicitly assumes that the base-model evolution $\Delta\Theta$ and the task-specific adaptation $\Delta\Theta_i$ can be combined with fixed unit weights. 
In practice, however, their relative importance may vary across downstream service tasks and update sources. 
The evolution delta may introduce useful general capabilities, irrelevant changes, or even conflicts with the existing adapter. 
Therefore, a more flexible strategy is needed to adaptively balance these two sources of knowledge.

To this end, we consider constructing an initialization by fusing the base-evolution knowledge and the old task-adaptation knowledge:
\begin{equation}
    \Theta_{\mathrm{init}} = \alpha \cdot \widetilde{\Delta\Theta} + \beta \cdot \Delta\Theta_i ,
\end{equation}
where $\alpha$ and $\beta$ control the contributions of the base-model evolution and the previously deployed task adapter, respectively. 
Here, $\widetilde{\Delta\Theta}$ denotes the representation of the base-evolution delta in a parameter space compatible with the LoRA adapter. 
For notational simplicity, we use $\Delta\Theta$ to denote this compatible representation when there is no ambiguity. 
The naive summation strategy can be regarded as a special case with $\alpha=1$ and $\beta=1$.

To validate the necessity of adaptive fusion, we conduct preliminary experiments on 20News~\cite{joachims1997probabilistic} and MMLU using LLaMA3.1-8B as the initial backbone $\Theta$. 
After further pretraining on OpenOrca~\cite{lian2023openorca}, we obtain the evolved backbone $\Theta'$. 
We first search for the optimal $\beta$ while fixing $\alpha=1$, and then search for the optimal $\alpha$ with the previously selected $\beta$. 
As shown in Figure~\ref{fig:beta} and Figure~\ref{fig:alpha}, the choice of fusion coefficients has a substantial impact on downstream service quality. 
On 20News, for example, increasing $\beta$ from 0.7 to 1.9 improves accuracy by 46.9\%. 
Given the selected $\beta$, varying $\alpha$ also leads to different performance, with optimal values smaller than 1 on both datasets. 
This indicates that the base-evolution delta should be incorporated selectively rather than naively added with full strength.

The comparison with naive fusion further confirms this point. 
The setting $(\alpha=1,\beta=1)$ achieves only 35.6\% accuracy on 20News and 52.4\% accuracy on MMLU. 
In contrast, by optimizing the scaling coefficients to $(\alpha=0.6,\beta=1.9)$ for 20News and $(\alpha=0.3,\beta=1.8)$ for MMLU, the accuracy increases to 77.2\% and 73.4\%, respectively. 
These findings demonstrate that a static fusion rule fails to fully exploit the knowledge contained in the old adapter and the base-model evolution. 
Therefore, an intelligent and adaptive strategy is needed to identify suitable fusion weights and construct a better starting point for re-adaptation.

Furthermore, even with a high-quality initialization, subsequent fine-tuning remains necessary to fully adapt the adapter to the downstream service task on the evolved backbone. 
However, conventional fine-tuning guided solely by the task loss may quickly drift away from this favorable initialization, slowing convergence and increasing time-to-readiness. 
A natural solution is to use regularization to constrain the optimization trajectory around the initialized adapter. 
Strong regularization can keep the model in a promising parameter region and accelerate early service-quality recovery. 
Nevertheless, if the regularization remains too strong throughout training, it may restrict task-specific refinement and harm the final performance.

This trade-off motivates a scheduled regularization strategy. 
As shown in Figure~\ref{fig:regularization}, strong regularization accelerates early convergence, but may constrain the final achievable accuracy because it limits exploration. 
In contrast, no regularization allows the model to explore more freely and may eventually reach a better local optimum, but it requires more steps to reach a high-quality region. 
Therefore, an effective re-adaptation process should first apply strong regularization to rapidly recover service-ready performance, and then relax the constraint to enable task-specific refinement. 
This observation motivates the two-stage design of \method, which combines adaptive LoRA initialization with scheduled regularization to achieve both fast time-to-readiness and high final service quality.

\vspace{-3mm}
\section{Method}\label{sec:algorithm}

\subsection{Overview}
\label{sec:method_overview}

Our proposed \method\ is designed to rapidly restore service-ready adapters after base-service evolution.
When the LLM backbone evolves from $\Theta$ to $\Theta'$, where $\Theta'=\Theta+\Delta\Theta$, previously deployed task adapters may no longer be fully compatible with the updated backbone.
Instead of retraining each adapter from scratch, \method\ reuses two forms of existing knowledge: the task-specific knowledge encoded in the previously deployed adapter $\Delta\Theta_i$, and the evolution knowledge reflected by the base-model parameter delta $\Delta\Theta$.
The goal is to efficiently construct an updated adapter for the evolved backbone such that the downstream service can quickly reach its required quality threshold.

As illustrated in Figure~\ref{fig: overview}, \method\ consists of two key steps.
\textcircled{1} \textbf{Adaptive LoRA initialization.}
This step constructs a compatibility-aware starting point, denoted by $\Theta_{\mathrm{init}}$, for re-adapting the $i$-th downstream service adapter on the evolved backbone.
Here, $\Theta_{\mathrm{init}}$ denotes the initialized LoRA update rather than the full backbone.
It is obtained by adaptively fusing the previously deployed task adapter $\Delta\Theta_i$ with the base-evolution delta $\Delta\Theta$, after mapping the latter into a compatible LoRA parameter space.
The fusion coefficients are selected through Bayesian optimization to reduce the number of pre-rollout validation probes.
\textcircled{2} \textbf{Fine-tuning with scheduled regularization.}
Starting from $\Theta_{\mathrm{init}}$, this step performs re-adaptation under a two-stage regularization schedule.
The first stage strongly anchors the adapter to the initialized state for fast service-readiness recovery, while the second stage relaxes the constraint to enable task-specific refinement.
Together, these two steps reduce time-to-readiness while preserving downstream service quality.

\subsection{Adaptive LoRA Initialization}
\label{sec:adaptive_initialization}

The purpose of adaptive LoRA initialization is to construct a high-quality starting point for re-adapting the old task adapter to the evolved backbone.
A naive strategy directly reuses $\Delta\Theta_i$ on $\Theta'$, while training from scratch ignores the knowledge already encoded in the deployed adapter.
Both strategies are suboptimal for fast service rollout.
In contrast, \method\ initializes the updated adapter by jointly exploiting the old task adaptation and the base-model evolution.

Before fusion, the base-evolution delta is mapped to the same LoRA parameter space as the task adapter.
This mapping ensures that both components are defined over the same target modules and have compatible ranks and parameterization.
We denote this mapping by $\Pi_r(\cdot)$, where $r$ is the rank used for downstream LoRA adaptation.
For each target module, $\Pi_r(\Delta\Theta)$ represents the rank-compatible LoRA update induced by the base-model evolution.
The initialized adapter is then formulated as:
\begin{equation}
\label{eq:init_formula}
    \Theta_{\mathrm{init}}(\alpha,\beta)
    =
    \alpha \cdot \Pi_r(\Delta\Theta)
    +
    \beta \cdot \Delta\Theta_i ,
\end{equation}
where $\alpha$ and $\beta$ are scalar coefficients that control the contribution of the base-evolution knowledge and the previously deployed task-specific knowledge, respectively.

The central challenge is to efficiently determine the optimal coefficients $(\alpha^*,\beta^*)$.
As shown in Section~\ref{subsec: motivation}, the quality of the initialization is sensitive to the choice of $(\alpha,\beta)$.
Although one-dimensional coefficient sweeps reveal informative trends, the joint optimum remains task-dependent and difficult to predict because it depends on the interaction between base-model evolution and task-specific adaptation.
A brute-force grid search is unattractive in service deployment.
A coarse grid may miss a high-quality initialization, while a fine-grained grid introduces excessive pre-rollout overhead, especially when many downstream service adapters need to be updated.

Therefore, \method\ adopts Bayesian Optimization (BayesOpt) to search for $(\alpha,\beta)$ in a sample-efficient manner.
BayesOpt is suitable here because directly evaluating the quality of a coefficient pair is expensive, and the relationship between $(\alpha,\beta)$ and adapter quality is a black-box function without analytical gradients~\cite{frazier2018tutorial}.
Since each validation probe delays service rollout, the search process should minimize the number of evaluations.
Following prior practice~\cite{liu2024checkpoint}, BayesOpt iteratively builds a probabilistic surrogate model for the objective function and selects the next coefficient pair using an acquisition function.

For a given coefficient pair $(\alpha,\beta)$, we define the objective function $f(\alpha,\beta)$ as the negative validation loss of the evolved backbone equipped with the initialized adapter:
\begin{equation}
\label{eq:objective}
    f(\alpha,\beta)
    =
    -\mathcal{L}_{\mathrm{val}}
    \bigl(\Theta', \Theta_{\mathrm{init}}(\alpha,\beta); D_{\mathrm{val}}\bigr),
\end{equation}
where $D_{\mathrm{val}}$ is a small held-out validation set sampled from the corresponding downstream service task.
Maximizing $f(\alpha,\beta)$ therefore favors an initialization that gives lower validation loss before full re-adaptation.
The optimal coefficients are obtained by:
\begin{equation}
\label{eq:max_task}
    (\alpha^*,\beta^*)
    =
    \arg\max_{\alpha,\beta} f(\alpha,\beta).
\end{equation}

We use a Gaussian Process (GP)~\cite{seeger2004gaussian} as the surrogate model.
Let $\mathbf{x}=(\alpha,\beta)$ denote a coefficient pair.
Given a set of observed coefficient pairs $\mathbf{X}_{1:N}=[\mathbf{x}_1,\ldots,\mathbf{x}_N]$ and their objective values $f(\mathbf{X}_{1:N})=[f(\mathbf{x}_1),\ldots,f(\mathbf{x}_N)]$, the GP prior is defined as:
\begin{equation}
\label{eq:gp_prior}
    f(\mathbf{X}_{1:N})
    \sim
    \mathcal{N}
    \bigl(
    \mu_0(\mathbf{X}_{1:N}),
    \Sigma_0(\mathbf{X}_{1:N},\mathbf{X}_{1:N})
    \bigr),
\end{equation}
where $\mu_0(\cdot)$ and $\Sigma_0(\cdot,\cdot)$ are the prior mean and covariance functions, respectively.
For a new candidate $\mathbf{x}_{N+1}$, the posterior distribution of $f(\mathbf{x}_{N+1})$ is:
\begin{equation}
\label{eq:gp_posterior}
    f(\mathbf{x}_{N+1}) \mid f(\mathbf{X}_{1:N})
    \sim
    \mathcal{N}
    \bigl(
    \mu_N(\mathbf{x}_{N+1}),
    \sigma_N^2(\mathbf{x}_{N+1})
    \bigr),
\end{equation}
where the posterior mean and variance are computed as:
\begin{align}
\label{eq:gp_mean}
    \mu_N(\mathbf{x}_{N+1})
    &=
    \Sigma_0(\mathbf{x}_{N+1},\mathbf{X}_{1:N})
    \Sigma_0^{-1}(\mathbf{X}_{1:N},\mathbf{X}_{1:N})
    \nonumber \\
    &\quad \cdot
    \bigl(
    f(\mathbf{X}_{1:N})-\mu_0(\mathbf{X}_{1:N})
    \bigr)
    +
    \mu_0(\mathbf{x}_{N+1}), \\
\label{eq:gp_var}
    \sigma_N^2(\mathbf{x}_{N+1})
    &=
    \Sigma_0(\mathbf{x}_{N+1},\mathbf{x}_{N+1})
    -
    \Sigma_0(\mathbf{x}_{N+1},\mathbf{X}_{1:N})
    \nonumber \\
    &\quad \cdot
    \Sigma_0^{-1}(\mathbf{X}_{1:N},\mathbf{X}_{1:N})
    \Sigma_0(\mathbf{X}_{1:N},\mathbf{x}_{N+1}).
\end{align}
We adopt the Squared Exponential, \ie, the RBF kernel:
\begin{equation}
\label{eq:kernel}
    \Sigma_0(\mathbf{x},\mathbf{x}')
    =
    \sigma_f^2
    \exp
    \left(
    -\frac{\|\mathbf{x}-\mathbf{x}'\|^2}{2l^2}
    \right),
\end{equation}
where $\sigma_f^2$ is the signal variance and $l$ is the length-scale.
These hyperparameters are estimated from the observed data during the search.

To select the next coefficient pair, we use Expected Improvement (EI) as the acquisition function.
EI balances exploitation, \ie, selecting regions with high predicted objective values, and exploration, \ie, selecting regions with high uncertainty.
It is defined as:
\begin{equation}
\label{eq:ei}
    EI(\mathbf{x})
    =
    \bigl(\mu_N(\mathbf{x})-f^+-\xi\bigr)\Phi(Z)
    +
    \sigma_N(\mathbf{x})\phi(Z),
\end{equation}
where $f^+$ is the best observed objective value, $\xi$ is a trade-off parameter, and
\begin{equation}
\label{eq:z_score}
    Z
    =
    \frac{\mu_N(\mathbf{x})-f^+-\xi}{\sigma_N(\mathbf{x})}.
\end{equation}
Here, $\Phi(\cdot)$ and $\phi(\cdot)$ denote the Cumulative Distribution Function (CDF) and Probability Density Function (PDF) of the standard normal distribution, respectively.
The next coefficient pair is selected by maximizing EI:
\begin{equation}
\label{eq:max_acquisition}
    (\alpha_{N+1},\beta_{N+1})
    =
    \arg\max_{\alpha,\beta} EI(\alpha,\beta).
\end{equation}
After $T$ search iterations, the best observed coefficient pair is used to construct $\Theta_{\mathrm{init}}^*$ according to Eq.~\eqref{eq:init_formula}.

\begin{algorithm}[!t]
\setstretch{0.9}
\caption{Adaptive LoRA initialization and fine-tuning with scheduled regularization}
\label{alg:relora}
\KwIn{Old backbone $\Theta$, evolved backbone $\Theta'$, old adapter $\Delta\Theta_i$, evolution delta $\Delta\Theta$, training set $D_{\mathrm{train}}$, validation set $D_{\mathrm{val}}$, search budget $T$, stage lengths $E_1$ and $E_{\mathrm{total}}$.}
\KwOut{Updated adapter $\Delta\Theta'_i$ for the evolved backbone.}

Initialize observation history $\mathcal{H}_0$ with a small set of evaluated coefficient pairs\;

\For{$k=0$ \KwTo $T-1$}{
    Fit the GP surrogate using the current observation history $\mathcal{H}_k$\;
    Compute the acquisition function $EI(\alpha,\beta)$ according to Eq.~\eqref{eq:ei}\;
    Select $(\alpha_{k+1},\beta_{k+1})$ by maximizing $EI(\alpha,\beta)$ according to Eq.~\eqref{eq:max_acquisition}\;
    Construct candidate initialized adapter
    $\Theta_{\mathrm{init}}(\alpha_{k+1},\beta_{k+1})
    =
    \alpha_{k+1}\Pi_r(\Delta\Theta)
    +
    \beta_{k+1}\Delta\Theta_i$\;
    Evaluate $f(\alpha_{k+1},\beta_{k+1})$ on $D_{\mathrm{val}}$ using $\Theta'$ with $\Theta_{\mathrm{init}}(\alpha_{k+1},\beta_{k+1})$\;
    Update the observation history
    $\mathcal{H}_{k+1}
    \leftarrow
    \mathcal{H}_k
    \cup
    \{((\alpha_{k+1},\beta_{k+1}), f(\alpha_{k+1},\beta_{k+1}))\}$\;
}

Select the best coefficients $(\alpha^*,\beta^*)$ according to Eq.~\eqref{eq:max_task}\;
Obtain $\Theta_{\mathrm{init}}^*$ by Eq.~\eqref{eq:init_formula}, and initialize $(A^{(0)},B^{(0)}) \leftarrow (A_{\mathrm{init}},B_{\mathrm{init}})$\;

\For{$t=0$ \KwTo $E_1-1$}{
    Sample a mini-batch $D^{(t)}$ from $D_{\mathrm{train}}$\;
    Compute task loss $\mathcal{L}_{\mathrm{task}}^{(t)}$ using $(A^{(t)},B^{(t)})$ on $D^{(t)}$\;
    Compute the stage-1 loss $\mathcal{L}_{S1}^{(t)}$ according to Eq.~\eqref{eq:loss_stage_1}\;
    Update $A^{(t+1)}$ and $B^{(t+1)}$ using learning rate $\eta_1$\;
}

\For{$t=E_1$ \KwTo $E_{\mathrm{total}}-1$}{
    Sample a mini-batch $D^{(t)}$ from $D_{\mathrm{train}}$\;
    Compute task loss $\mathcal{L}_{\mathrm{task}}^{(t)}$ using $(A^{(t)},B^{(t)})$ on $D^{(t)}$\;
    Compute the stage-2 loss $\mathcal{L}_{S2}^{(t)}$ according to Eq.~\eqref{eq:loss_stage_2}\;
    Compute the annealed learning rate $\eta(t)$ according to Eq.~\eqref{eq:lr_cosine}\;
    Update $A^{(t+1)}$ and $B^{(t+1)}$ using learning rate $\eta(t)$\;
}

Return the final adapter $\Delta\Theta'_i=(A^{(E_{\mathrm{total}})},B^{(E_{\mathrm{total}})})$\;
\end{algorithm}

\subsection{Fine-tuning with Scheduled Regularization}
\label{sec:scheduled_regularization}

After obtaining the optimized initialization $\Theta_{\mathrm{init}}^*$, \method\ further re-adapts the adapter on the downstream task training set $D_{\mathrm{train}}$.
Although a compatibility-aware initialization provides a favorable starting point, directly fine-tuning with only the task loss may cause the adapter to drift away from this initialization too quickly, which can slow convergence and increase time-to-readiness.
As observed in Figure~\ref{fig:regularization}, strong regularization accelerates early convergence but may restrict final refinement, whereas no regularization provides greater flexibility but requires more steps to reach a high-quality region.
To balance these two effects, \method\ adopts a scheduled regularization strategy with two stages: fast service-readiness recovery and quality refinement.

For clarity, the following equations are written for one adapted module, and the same procedure is applied to all LoRA-adapted modules.
Let $\Theta_{\mathrm{init}}^*=(A_{\mathrm{init}},B_{\mathrm{init}})$ denote the optimized initialized adapter, and let $\Delta\Theta_i^{(t)}=(A^{(t)},B^{(t)})$ denote the trainable LoRA factors at step $t$.

\subsubsection{Stage 1: Guided Rapid Adaptation for Service-Readiness Recovery}
\label{sec:scheduled_regularization_stage1}

The objective of the first stage is to rapidly move the adapter into a service-ready quality region.
To this end, \method\ anchors the optimization trajectory around the initialized adapter by adding strong L2 regularization to the task loss.
At training step $t$, the stage-1 objective is:
\begin{equation}
\label{eq:loss_stage_1}
    \mathcal{L}_{S1}^{(t)}
    =
    \mathcal{L}_{\mathrm{task}}^{(t)}
    +
    \frac{\lambda_1}{2}
    \left(
    \|A^{(t)}-A_{\mathrm{init}}\|_F^2
    +
    \|B^{(t)}-B_{\mathrm{init}}\|_F^2
    \right),
\end{equation}
where $\|\cdot\|_F$ denotes the Frobenius norm, and $\lambda_1$ controls the strength of the anchor.
The gradient with respect to $A^{(t)}$ is:
\begin{equation}
\label{eq:gradient_A_stage_1}
    \nabla_{A^{(t)}}\mathcal{L}_{S1}^{(t)}
    =
    \nabla_{A^{(t)}}\mathcal{L}_{\mathrm{task}}^{(t)}
    +
    \lambda_1(A^{(t)}-A_{\mathrm{init}}),
\end{equation}
with the gradient for $B^{(t)}$ computed analogously.
Given learning rate $\eta_1$, the update for $A^{(t)}$ is:
\begin{equation}
\label{eq:update_A_stage_1}
    A^{(t+1)}
    =
    A^{(t)}
    -
    \eta_1
    \nabla_{A^{(t)}}\mathcal{L}_{S1}^{(t)},
\end{equation}
and $B^{(t)}$ is updated in the same manner.

This strong anchoring encourages the adapter to stay close to $\Theta_{\mathrm{init}}^*$, which already integrates knowledge from the base-model evolution and the previously deployed task adapter.
By using a relatively large $\lambda_1$ during the first $E_1$ steps, \method\ reduces unnecessary exploration in the early phase and accelerates service-quality recovery.

\subsubsection{Stage 2: Refinement and Exploration for Quality Improvement}
\label{sec:scheduled_regularization_stage2}

After the adapter reaches a promising parameter region, the second stage relaxes the anchor to enable more task-specific refinement.
The regularization strength is reduced to $\lambda_2$, where $\lambda_2 \ll \lambda_1$.
The stage-2 objective becomes:
\begin{equation}
\label{eq:loss_stage_2}
    \mathcal{L}_{S2}^{(t)}
    =
    \mathcal{L}_{\mathrm{task}}^{(t)}
    +
    \frac{\lambda_2}{2}
    \left(
    \|A^{(t)}-A_{\mathrm{init}}\|_F^2
    +
    \|B^{(t)}-B_{\mathrm{init}}\|_F^2
    \right).
\end{equation}
With a weaker anchor, the optimization is primarily driven by the task loss while still retaining mild protection against catastrophic drift from the initialization.
This allows the adapter to refine its task-specific behavior and approach a better local optimum.

Meanwhile, a cosine annealing learning-rate schedule~\cite{loshchilov2016sgdr} is applied during the second stage:
\begin{equation}
\label{eq:lr_cosine}
    \eta(t)
    =
    \eta_{\min}
    +
    \frac{1}{2}(\eta_1-\eta_{\min})
    \left(
    1+
    \cos
    \left(
    \frac{(t-E_1)\pi}{E_{\mathrm{total}}-E_1}
    \right)
    \right),
\end{equation}
where $\eta_{\min}$ is the minimum learning rate, $\eta_1$ is the initial learning rate of the second stage, and $E_{\mathrm{total}}$ is the total number of re-adaptation steps across both stages.
This schedule gradually reduces the step size, enabling stable refinement after the rapid recovery phase.
The final adapter after $E_{\mathrm{total}}$ steps is used as the updated task adapter $\Delta\Theta'_i$.

\vspace{-3mm}
\section{Evaluation}\label{sec:evaluation}

\subsection{Experimental Setup}
\label{sec:experimental_setup}

All experiments described in this paper are conducted on a single server equipped with an Intel(R) Xeon(R) Platinum 8358P CPU, 8 NVIDIA GeForce RTX A6000 GPUs (each with 48 GB of memory), and 512 GB of RAM.

\textbf{Models.}
We evaluate \method\ across multiple LLM service backbones to demonstrate its generality, including LLaMA2-7B~\cite{touvron2023llama}, LLaMA3.1-8B~\cite{grattafiori2024llama}, and Mistral-7B~\cite{chaplot2023albert}.
These models cover different architectures and model families, allowing us to examine whether \method\ remains effective under heterogeneous service backbones.

\textbf{Datasets and Service Tasks.}
We evaluate \method\ on six datasets covering representative downstream service tasks.
The statistics of these datasets are summarized in Table~\ref{tab:datasets}.
Specifically, we consider the following service categories.
\begin{itemize}
    \item \textit{Knowledge and Reasoning Service}: We use the MMLU benchmark~\cite{hendrycks2020measuring}, which contains questions across 57 subjects, to evaluate general reasoning capability.
    
    \item \textit{Sentiment Analysis Service}: We use SST-2~\cite{wang2018glue}, which consists of movie reviews annotated with binary labels.
    
    \item \textit{News and Topic Classification Services}: We use AGNews~\cite{zhang2015character}, a four-class news classification dataset, and 20News~\cite{joachims1997probabilistic} containing documents from 20 categories.
    
    \item \textit{Natural Language Inference Service}: We use MNLI~\cite{wang2018glue} and SNLI~\cite{bowman2015large} to evaluate the ability to determine logical relationships between sentence pairs.
\end{itemize}

\begin{table}[!t]
\caption{Statistics of downstream service tasks in experiments.}
\vspace{0.1cm}
\centering
\resizebox{\columnwidth}{!}{
\begin{tabular}{l|cccccc}
\hline
Service Type 
& Sentiment 
& News 
& Topic 
& NLI 
& NLI 
& Knowledge \\
& Analysis 
& Classification 
& Classification 
& (Multi) 
& (Single) 
& \& Reasoning \\
\hline
Dataset & SST-2 & AGNews & 20News & MNLI & SNLI & MMLU \\
\hline
\#Train & 67,349 & 120,000 & 11,314 & 392,702 & 550,152 & 20,000 \\
\#Test  & 1,821  & 7,600   & 7,532  & 9,815   & 20,000  & 2,000 \\
\hline
\end{tabular}}
\label{tab:datasets}
\end{table}

\textbf{Training Details.}
To simulate base-service evolution, we employ continual pretraining~\cite{qin2023recyclable,qin2022elle}.
Specifically, we start from a pretrained LLM backbone $\Theta$ and update it on a specific pretraining corpus to obtain an evolved backbone $\Theta'$.
The evolution process is implemented using LoRA~\cite{hu2021lora,khan2024portllm} with rank $r=64$, learning rate $2\text{e}^{-4}$, and 4 training epochs, since full-parameter updating is infeasible under our platform resource constraints.
The continued pretraining datasets include OpenOrca~\cite{lian2023openorca}, AlpacaGPT4~\cite{peng2023instruction}, and OpenPlatypus~\cite{lee2023platypus}, which represent different update sources for LLM services.
For downstream service adaptation, we use LoRA with rank $r=16$ and a uniform batch size of 32 for all methods.
For the adaptive initialization search in \method, we use a small validation set of 512 examples sampled from the corresponding downstream service task.
We train for 5 epochs on MMLU and 2 epochs for all other datasets.
\method\ uses a cosine annealing learning-rate schedule with $\eta_1=5\text{e}^{-4}$ and $\eta_{\min}=1\text{e}^{-5}$~\cite{zhang2023fedpetuning,yao2024ferrari}.
Based on the hyperparameter sensitivity analysis in Section~\ref{sec:robustness}, we set the regularization strengths to $\lambda_1=1\text{e}^{-3}$ in Stage 1 and $\lambda_2=1\text{e}^{-4}$ in Stage 2, which provides a balance between convergence speed and final service quality.

\textbf{Metrics.}
We evaluate service quality using accuracy on the test set of each downstream service task.
For the original backbone $\Theta$ and the evolved backbone $\Theta'$, we report zero-shot accuracy.
For task-adaptation methods, we report the accuracy after adapter adaptation.
To assess rollout efficiency, we use two time-related metrics.
First, \textit{time-to-readiness} measures the wall-clock time required for an updated adapter to reach a predefined service-quality threshold after the base model evolves.
This metric is reported in seconds.
Second, \textit{end-to-end service rollout time} measures the total time required by the complete update process, including search and re-adaptation when applicable.
This metric is reported in minutes.
For fairness, the target service-quality threshold is set to a level that all methods can reach.

\textbf{Baselines.}
We compare \method\ with the following baselines.
Under zero-shot settings, we include:
(1) \textit{Pretrained LLM $\Theta$}, the original backbone before service evolution; and
(2) \textit{Evolved LLM $\Theta'$}, the updated backbone after continual pretraining.
These two baselines help quantify the effect of base-model evolution without task-specific adaptation.

For task-adaptation methods, we include:
(3) \textit{LoRA-Scratch}, which reconstructs a new LoRA adapter from scratch on $\Theta'$ for each downstream task;
(4) \textit{PortLLM+FT}~\cite{khan2024portllm}, which performs naive adapter transfer followed by fine-tuning; and
(5) \textit{ORAL}~\cite{khan2025oral}, a conditional recurrent diffusion framework that generates LoRA weights from scratch for the evolved backbone.
Unless otherwise specified, the training cost of ORAL's generative model is not included.
Including ORAL enables a comparison between re-adapting an existing deployed adapter and generating a new adapter from task specifications.

\begin{table*}[ht]
\centering
\caption{Service-quality comparison under different base-service update sources.}
\label{tab:performance_lLaMA}
\vspace{0.5mm}
\scalebox{0.85}{
\begin{tabular}{llc|ccccccc}
\toprule
\textbf{Update Source} & \multicolumn{2}{c}{\textbf{Baselines}} 
& \textbf{MMLU} & \textbf{SST-2} & \textbf{AGNews} & \textbf{20News} & \textbf{MNLI} & \textbf{SNLI} & \textbf{Avg. (↑)} \\
\midrule
& Zero-shot & Pretrained LLM $\Theta$ 
& 26.2 & 56.9 & 28.4 & 4.6 & 31.4 & 30.2 & 29.6 \\
\midrule
\multirow{5}{*}{OpenOrca} 
& Zero-shot & Evolved LLM $\Theta'$ 
& 24.5 & 56.0 & 29.2 & 3.7 & 32.3 & 33.0 & 29.8 \\
\cmidrule(l){2-10}
& \multirow{4}{*}{Task-adaptation Methods} 
& LoRA-Scratch 
& 81.4 & 96.3 & 95.2 & 76.2 & 91.6 & 91.8 & 88.8 \\
& & ORAL 
& 81.2 & 96.4 & 95.0 & 76.1 & 91.9 & 92.0 & 88.8 \\
& & PortLLM+FT 
& 82.5 & 96.6 & 95.2 & 76.4 & 92.6 & 92.1 & 89.2 \\
\cmidrule(l){4-10}
& & \method\ (Ours) 
& \textbf{83.9} & \textbf{97.0} & \textbf{95.3} & \textbf{76.6} & \textbf{93.5} & \textbf{93.1} & \textbf{89.9} \\
\midrule
\multirow{5}{*}{AlpacaGPT4} 
& Zero-shot & Evolved LLM $\Theta'$ 
& 24.0 & 51.6 & 31.0 & 4.5 & 31.6 & 31.4 & 29.0 \\
\cmidrule(l){2-10}
& \multirow{4}{*}{Task-adaptation Methods} 
& LoRA-Scratch 
& 79.5 & 96.1 & 95.0 & 75.6 & 91.5 & 92.1 & 88.3 \\
& & ORAL 
& 79.7 & 96.0 & 95.1 & 75.5 & 91.8 & 92.3 & 88.4 \\
& & PortLLM+FT 
& 81.7 & 96.4 & 95.2 & 76.1 & 92.4 & 92.7 & 89.1 \\
\cmidrule(l){4-10}
& & \method\ (Ours) 
& \textbf{82.9} & \textbf{96.9} & \textbf{95.4} & \textbf{76.3} & \textbf{93.0} & \textbf{93.1} & \textbf{89.6} \\
\midrule
\multirow{5}{*}{OpenPlatypus} 
& Zero-shot & Evolved LLM $\Theta'$ 
& 24.0 & 52.2 & 29.9 & 3.8 & 31.9 & 30.5 & 28.7 \\
\cmidrule(l){2-10}
& \multirow{4}{*}{Task-adaptation Methods} 
& LoRA-Scratch 
& 79.7 & 96.1 & 95.2 & 74.3 & 91.6 & 92.2 & 88.2 \\
& & ORAL 
& 79.4 & 96.2 & 95.0 & 74.6 & 91.4 & 92.4 & 88.2 \\
& & PortLLM+FT 
& 82.8 & 96.4 & 95.2 & 75.8 & 92.7 & 93.0 & 89.3 \\
\cmidrule(l){4-10}
& & \method\ (Ours) 
& \textbf{84.3} & \textbf{97.0} & \textbf{95.3} & \textbf{76.4} & \textbf{93.3} & \textbf{93.1} & \textbf{89.9} \\
\bottomrule
\end{tabular}}
\end{table*}

\begin{table*}[ht]
\centering
\caption{Service-quality recovery across different LLM service backbones.}
\label{tab:performance_models}
\vspace{0.5mm}
\scalebox{0.85}{
\begin{tabular}{llc|ccccccc}
\toprule
\textbf{Models} & \multicolumn{2}{c}{\textbf{Baselines}} 
& \textbf{MMLU} & \textbf{SST-2} & \textbf{AGNews} & \textbf{20News} & \textbf{MNLI} & \textbf{SNLI} & \textbf{Avg. (↑)} \\
\midrule
\multirow{5}{*}{LLaMA2-7B}
& Zero-shot & Evolved LLM $\Theta'$ 
& 25.0 & 52.0 & 26.3 & 5.1 & 33.3 & 33.1 & 29.1 \\
\cmidrule(l){2-10}
& \multirow{4}{*}{Task-adaptation Methods}
& LoRA-Scratch 
& 67.9 & 96.6 & 95.3 & 75.5 & 90.7 & \textbf{91.8} & 86.3 \\
& & ORAL 
& 67.8 & 96.7 & 95.2 & 75.5 & 90.8 & 91.4 & 86.2 \\
& & PortLLM+FT 
& 68.3 & 97.1 & \textbf{95.4} & 75.7 & 90.7 & 91.6 & 86.5 \\
& & \method\ (Ours) 
& \textbf{70.9} & \textbf{97.5} & \textbf{95.4} & \textbf{75.8} & \textbf{91.0} & 91.7 & \textbf{87.1} \\
\midrule
\multirow{5}{*}{Mistral-7B}
& Zero-shot & Evolved LLM $\Theta'$ 
& 25.2 & 50.9 & 24.2 & 6.2 & 33.3 & 33.3 & 28.9 \\
\cmidrule(l){2-10}
& \multirow{4}{*}{Task-adaptation Methods}
& LoRA-Scratch 
& 75.3 & 96.8 & 95.3 & 73.4 & \textbf{91.5} & \textbf{92.1} & 87.4 \\
& & ORAL 
& 75.5 & 96.7 & 95.2 & 72.9 & 90.8 & 91.6 & 87.1 \\
& & PortLLM+FT 
& 76.0 & 96.6 & 95.3 & 74.9 & 91.0 & 91.9 & 87.6 \\
& & \method\ (Ours) 
& \textbf{76.6} & \textbf{96.9} & \textbf{95.5} & \textbf{76.3} & 91.2 & \textbf{92.1} & \textbf{88.1} \\
\midrule
\multirow{5}{*}{LLaMA3.1-8B}
& Zero-shot & Evolved LLM $\Theta'$ 
& 24.5 & 56.0 & 29.2 & 3.7 & 32.3 & 33.0 & 29.8 \\
\cmidrule(l){2-10}
& \multirow{4}{*}{Task-adaptation Methods}
& LoRA-Scratch 
& 81.4 & 96.3 & 95.2 & 76.2 & 91.6 & 91.8 & 88.8 \\
& & ORAL 
& 81.2 & 96.4 & 95.0 & 76.1 & 91.9 & 92.0 & 88.8 \\
& & PortLLM+FT 
& 82.5 & 96.6 & 95.2 & 76.4 & 92.6 & 92.1 & 89.2 \\
& & \method\ (Ours) 
& \textbf{83.9} & \textbf{97.0} & \textbf{95.3} & \textbf{76.6} & \textbf{93.5} & \textbf{93.1} & \textbf{89.9} \\
\bottomrule
\end{tabular}}
\end{table*}

\begin{figure}[!t]
\centering
\includegraphics[width=0.4\textwidth]{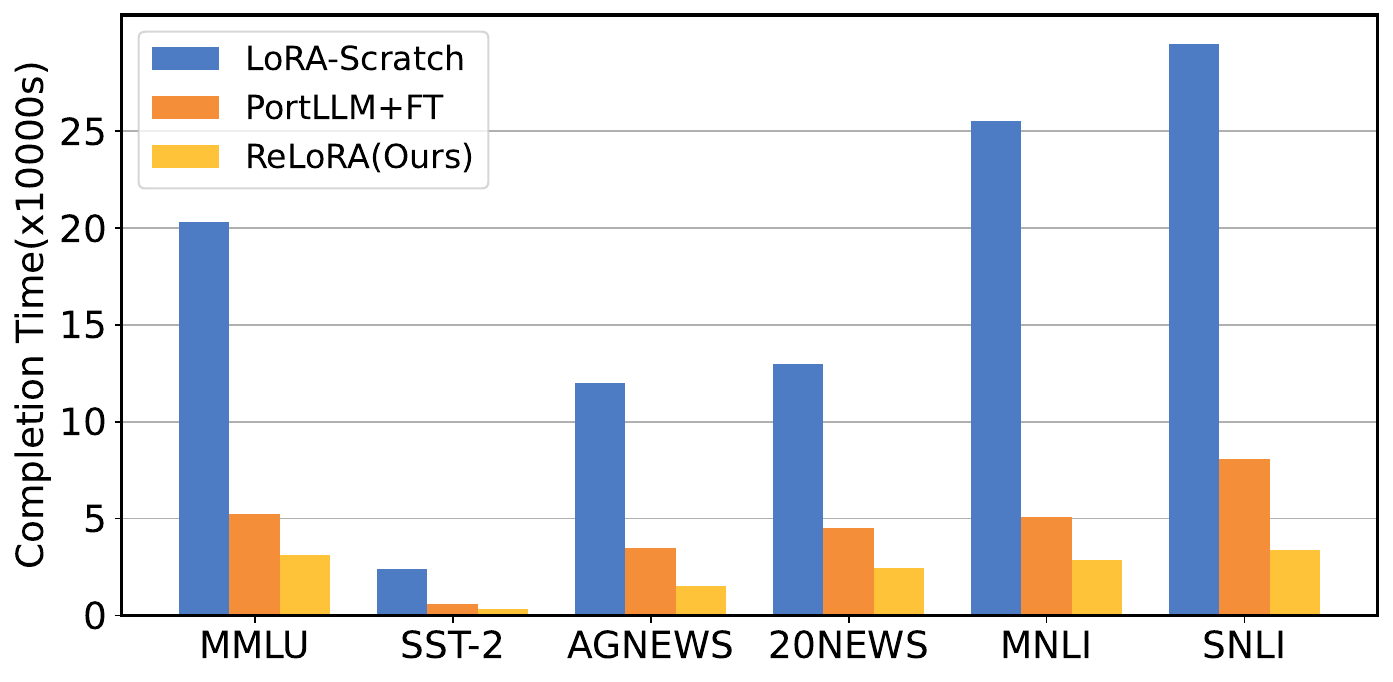}
\caption{Time-to-readiness of different methods to reach the target service-quality threshold across downstream tasks.}
\label{fig:speed_up}
\end{figure}

\subsection{Overall Performance}
\label{sec:overall_performance}

We first evaluate the overall effectiveness of \method\ against the baselines.
The service-quality results are summarized in Tables~\ref{tab:performance_lLaMA} and~\ref{tab:performance_models}.
We make four key observations.

First, compared with the evolved backbone $\Theta'$, \method\ consistently improves service quality across all downstream tasks.
For example, when LLaMA3.1-8B is further pretrained on OpenOrca, \method\ improves the service quality on \{MMLU, SST-2, AGNews, 20News, MNLI, SNLI\} by \{59.4\%, 41.0\%, 66.1\%, 72.9\%, 61.2\%, 60.1\%\}, respectively.
These results show that task-specific adaptation remains necessary even after the base model has been updated through continued pretraining.

Second, \method\ improves not only final service quality but also rollout efficiency.
As shown in Figure~\ref{fig:speed_up}, we report the time-to-readiness required by each method to reach a predefined target service-quality threshold.
The target thresholds for \{MMLU, SST-2, AGNews, 20News, MNLI, SNLI\} are set to \{81\%, 96\%, 95\%, 75\%, 91\%, 92\%\}, respectively.
All experiments in Figure~\ref{fig:speed_up} are conducted using LLaMA3.1-8B with OpenOrca as the update source.
On MNLI, for instance, \method\ reaches 91\% service quality in 28,802 seconds, whereas LoRA-Scratch requires 255,328 seconds.
This corresponds to an 8.9$\times$ speedup and an 88.7\% reduction in time-to-readiness.
Meanwhile, \method\ also improves final accuracy over LoRA-Scratch by 1.9\% and 2.5\% on MNLI and MMLU, respectively.
These results demonstrate that \method\ can accelerate service readiness while maintaining the final quality of the updated adapter.

Third, \method\ achieves clear advantages over PortLLM+FT in both convergence speed and final service quality.
For example, on MNLI, \method\ achieves a 4.0$\times$ speedup and improves service quality by 1.1\% compared with PortLLM+FT, which initializes fine-tuning by directly transferring the old adapter, whereas \method\ adaptively fuses the old task adapter with the base-model evolution delta.
This compatibility-aware initialization provides a better starting point for re-adaptation, resulting in faster convergence and higher service quality.

Fourth, \method\ outperforms ORAL, a representative generative adapter baseline.
As shown in Tables~\ref{tab:performance_lLaMA} and~\ref{tab:performance_models}, ORAL generally achieves service quality comparable to LoRA-Scratch.
For example, on LLaMA3.1-8B with OpenOrca, ORAL and LoRA-Scratch both achieve an average service quality of 88.8\%.
This confirms that ORAL can generate task-specific adapters for evolved backbones.
However, \method\ consistently obtains better performance by reusing the fine-grained task knowledge encoded in the previously deployed adapter.
On the same setting, \method\ achieves an average service quality of 89.9\%, outperforming ORAL by 1.1\%.
The advantage is more pronounced on MMLU, where \method\ reaches 83.9\%, exceeding ORAL by 2.7\%.
These results indicate that re-adapting an existing specialized adapter can be more effective than generating a new adapter from general task descriptions.

\subsection{Optimization Analysis and System Efficiency}
\label{sec:system_efficiency}

\noindent\textbf{Efficiency of Optimization Strategy.}
A potential concern of adaptive initialization is the additional cost of searching for the fusion coefficients $\alpha$ and $\beta$.
To evaluate this cost, we compare the pre-rollout compatibility search overhead of the GP-based Bayesian optimization strategy with a standard grid search baseline.
A grid search with 10 values per coefficient requires 100 evaluations, whereas our GP-based search uses only $T=20$ evaluations.
Thus, the GP-based strategy reduces the number of evaluations by 5$\times$.

Table~\ref{tab:alg_overhead} empirically confirms this advantage.
For example, on MMLU, grid search consumes 20,786 seconds, which could offset the benefit of fast re-adaptation.
In contrast, the GP-based search used by \method\ takes only 4,573 seconds.
On smaller datasets such as SST-2, the search overhead is only 634 seconds.
These results show that the proposed search strategy provides a sample-efficient way to identify compatibility-aware initialization coefficients.

\begin{table}[t]
\centering
\caption{Pre-rollout compatibility search overhead.}
\label{tab:alg_overhead}
\vspace{0.1cm}
\resizebox{\columnwidth}{!}{
\begin{tabular}{l|cccccc}
\hline
Method & SST-2 & AGNews & 20News & SNLI & MNLI & MMLU \\
\hline
Grid search (s) & 2884 & 1748 & 13691 & 1421 & 2279 & 20786 \\
GP search (s)  & 634  & 383  & 3071  & 312  & 496  & 4573 \\
Reduction      & 78.0\% & 78.1\% & 77.6\% & 78.0\% & 78.2\% & 78.0\% \\
\hline
\end{tabular}}
\end{table}

\noindent\textbf{Analysis of Fusion Coefficients.}
To understand how \method\ balances base-model evolution and task-specific adaptation, we analyze the optimal fusion coefficients $(\alpha,\beta)$ identified under different downstream service tasks and update sources.
The results are summarized in Table~\ref{tab:best_alpha_beta}.

We observe that $\beta$ is consistently larger than 1.0, typically ranging from 1.4 to 1.9.
This suggests that the task-specific knowledge encoded in the previously deployed adapter remains highly valuable and often needs to be amplified to compensate for the representation shift induced by base-model evolution.
At the same time, $\alpha$ remains nonzero across all settings, ranging from 0.1 to 0.7.
This indicates that the base-evolution delta also provides useful information and should be incorporated into the initialization.
For complex reasoning tasks such as MMLU, $\alpha$ tends to be higher, reaching 0.7 under OpenOrca and OpenPlatypus.
This implies that general capability changes introduced by base-model evolution are especially important for knowledge-intensive tasks.
The variation of $(\alpha,\beta)$ across tasks and update sources further confirms that a static fusion rule cannot reliably handle heterogeneous service updates, motivating the adaptive search mechanism in \method.

\begin{table}[t]
\caption{Optimal service-specific fusion coefficients under different update sources.}
\label{tab:best_alpha_beta}
\vspace{0.5mm}
\centering
\resizebox{\columnwidth}{!}{
\begin{tabular}{l|cccccc}
\hline
Update Source & SST-2 & AGNews & 20News & SNLI & MNLI & MMLU \\
\hline
OpenOrca      & (0.1, 1.7) & (0.1, 1.5) & (0.3, 1.8) & (0.3, 1.5) & (0.4, 1.7) & (0.7, 1.9) \\
AlpacaGPT4    & (0.1, 1.8) & (0.1, 1.4) & (0.5, 1.9) & (0.3, 1.5) & (0.4, 1.6) & (0.6, 1.8) \\
OpenPlatypus  & (0.2, 1.6) & (0.1, 1.7) & (0.6, 1.7) & (0.4, 1.4) & (0.4, 1.4) & (0.7, 1.9) \\
\hline
\end{tabular}}
\vspace{-0.2cm}
\end{table}

\noindent\textbf{End-to-End Service Rollout Time.}
Beyond the time-to-readiness required to reach a target threshold, practical deployment also requires evaluating the total cost of the complete update process.
Table~\ref{tab:system_overhead} reports the end-to-end service rollout time on LLaMA3.1-8B with OpenOrca as the update source.
For \method, this time includes both the Bayesian optimization search and the re-adaptation.

Although \method\ introduces an additional search step, its total rollout time remains substantially lower than the baselines.
For example, on SNLI, \method\ completes the rollout process in 1,264 minutes, while LoRA-Scratch requires 2,813 minutes.
Aggregated over all six tasks, \method\ reduces the total rollout time from 12,628 minutes to 5,437 minutes, achieving a 56.9\% reduction and a 2.32$\times$ speedup over LoRA-Scratch.
These results show that the search overhead is well compensated by the reduction in re-adaptation time, making \method\ suitable for service maintenance.

\begin{table}[t]
\centering
\caption{End-to-end service rollout time including search and re-adaptation (minutes).}
\label{tab:system_overhead}
\vspace{1mm}
\resizebox{\columnwidth}{!}{
\begin{tabular}{l|cccccc|cc}
\hline
Method & SST-2 & AGNews & 20News & SNLI & MNLI & MMLU & Total Reduction & Total Speedup \\
\hline
LoRA-Scratch  & 567 & 1496 & 1672 & 2813 & 2791 & 3289 & -- & 1.00$\times$ \\
PortLLM+FT    & 321 & 879  & 955  & 1538 & 1331 & 1647 & 47.2\% & 1.89$\times$ \\
\method\ (Ours) & \textbf{273} & \textbf{746} & \textbf{828} & \textbf{1264} & \textbf{1009} & \textbf{1317} & \textbf{56.9\%} & \textbf{2.32$\times$} \\
\hline
\end{tabular}}
\end{table}

\begin{table}[t]
\centering
\caption{End-to-end service rollout efficiency derived from Table~\ref{tab:system_overhead}.}
\label{tab:rollout_efficiency}
\vspace{1mm}
\resizebox{\columnwidth}{!}{
\begin{tabular}{l|cccccc|c}
\hline
Dataset & SST-2 & AGNews & 20News & SNLI & MNLI & MMLU & Total \\
\hline
LoRA-Scratch (min) & 567 & 1496 & 1672 & 2813 & 2791 & 3289 & 12628 \\
\method\ (min)     & 273 & 746  & 828  & 1264 & 1009 & 1317 & 5437 \\
Reduction vs Scratch & 51.9\% & 50.1\% & 50.5\% & 55.1\% & 63.8\% & 60.0\% & 56.9\% \\
Speedup vs Scratch   & 2.08$\times$ & 2.01$\times$ & 2.02$\times$ & 2.23$\times$ & 2.77$\times$ & 2.50$\times$ & 2.32$\times$ \\
\hline
\end{tabular}}
\end{table}

Table~\ref{tab:rollout_efficiency} further breaks down the rollout efficiency by task.
The results highlight the practical service-maintenance benefit of \method: it not only accelerates training convergence, but also shortens the time required to bring updated downstream LLM services back online.

\subsection{Ablation Study}
\label{sec:ablation}

We further evaluate the contribution of the two core components of \method: adaptive LoRA initialization, denoted as AdaInit, and scheduled regularization, denoted as SReg.
The results are reported in Table~\ref{tab:ablation_lLaMA}.

To assess the effect of AdaInit, we compare \method\ without SReg against PortLLM+FT.
The former uses the adaptive initialization produced by Bayesian optimization but performs standard fine-tuning afterward, while the latter starts from naive adapter transfer.
The average service quality improves from 89.2\% to 89.5\%, indicating that compatibility-aware fusion provides a better starting point than direct adapter transfer.

To assess the effect of SReg, we compare full \method\ with \method\ without SReg.
Full \method\ improves the service quality from 89.5\% to 89.9\%.
This improvement, together with the faster convergence observed in previous experiments, shows that scheduled regularization helps the adapter move from the fused initialization to a better task-specific optimum.
Overall, both AdaInit and SReg are important for achieving fast and high-quality service recovery.

\begin{table}[!t]
\caption{Component contribution to service-quality recovery.}
\label{tab:ablation_lLaMA}
\vspace{1mm}
\centering
\resizebox{\columnwidth}{!}{
\begin{tabular}{l|cccccc|c}
\hline
Method & MMLU & SST-2 & AGNews & 20News & MNLI & SNLI & \textbf{Avg. (↑)} \\
\hline
\method\                  & \textbf{83.9} & \textbf{97.0} & \textbf{95.3} & \textbf{76.6} & \textbf{93.5} & \textbf{93.1} & \textbf{89.9} \\
\method\ w/o SReg         & 83.1 & 96.8 & 95.2 & 76.5 & 92.9 & 92.7 & 89.5 \\
PortLLM+FT                & 82.5 & 96.6 & 95.2 & 76.4 & 92.6 & 92.1 & 89.2 \\
\hline
\end{tabular}}
\end{table}

\subsection{Robustness Across Update Sources and Service Backbones}
\label{sec:robustness}

\begin{figure}[t]
\centering
\includegraphics[width=0.235\textwidth,height=3.2cm]{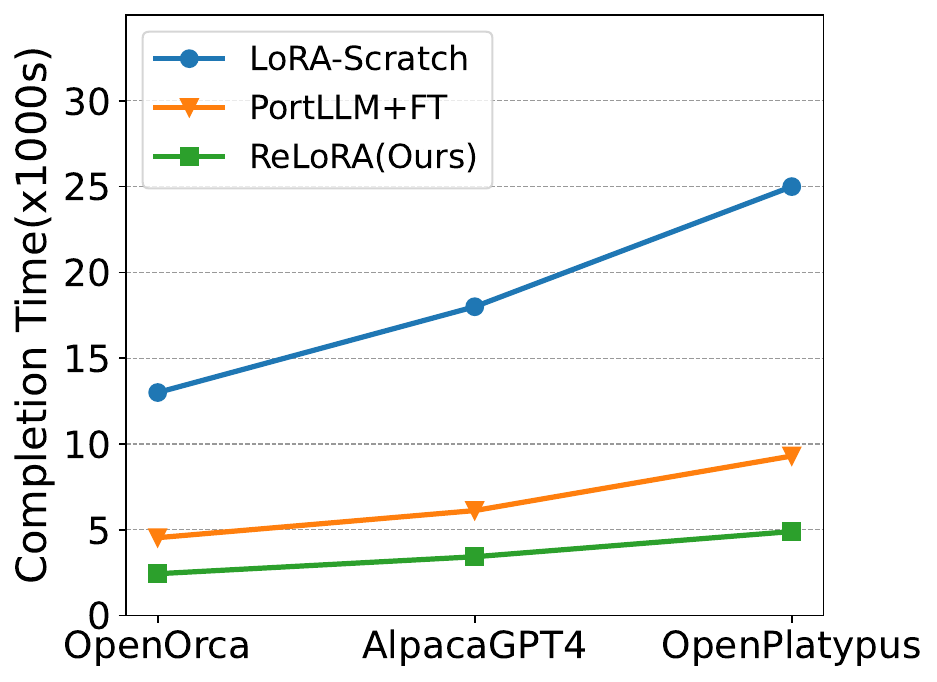}
\includegraphics[width=0.235\textwidth,height=3.2cm]{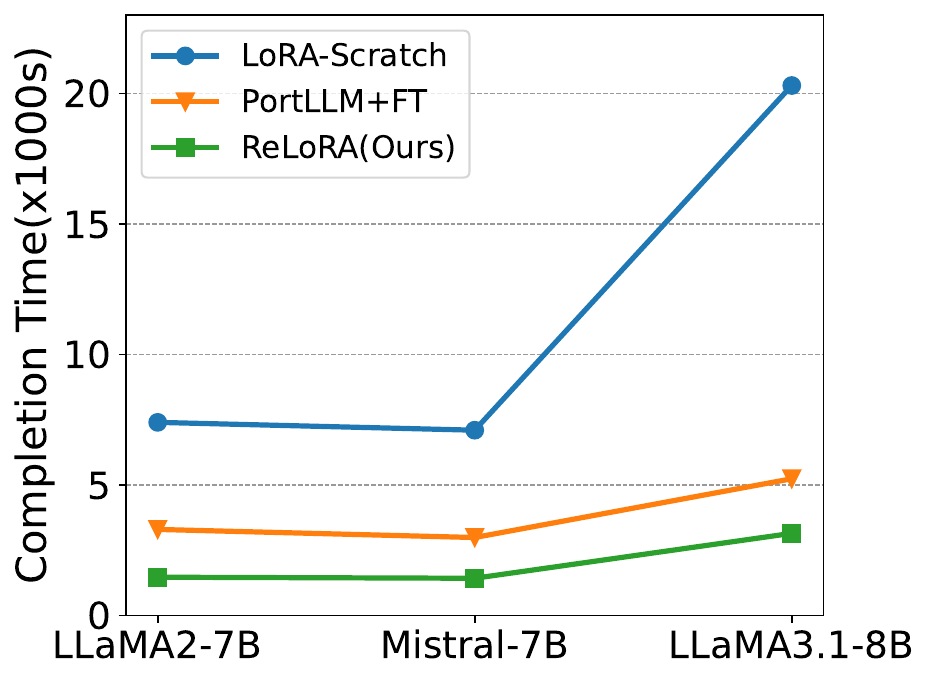}
\caption{Time-to-readiness under different update sources and LLM service backbones.}
\label{fig:completion_time_models_datasets}
\end{figure}

\noindent\textbf{Effect of Different Update Sources.}
We first evaluate whether \method\ remains robust under different base-service update sources.
Specifically, we use OpenOrca, AlpacaGPT4, and OpenPlatypus as continued pretraining datasets for LLaMA3.1-8B.
As shown in Table~\ref{tab:performance_lLaMA} and Figure~\ref{fig:completion_time_models_datasets}, \method\ consistently improves service quality and time-to-readiness across different update sources.

For example, on MNLI, \method\ improves accuracy over LoRA-Scratch by 1.9\%, 1.5\%, and 1.7\% under OpenOrca, AlpacaGPT4, and OpenPlatypus, respectively.
We also observe that different update sources affect the zero-shot performance of the evolved backbone.
For instance, continued pretraining on AlpacaGPT4 decreases zero-shot performance on MMLU and SST-2 compared with the original backbone $\Theta$.
This suggests that base-model evolution can introduce task-dependent shifts.
By incorporating the evolution delta into adapter initialization, \method\ mitigates these shifts and achieves more reliable service-quality recovery.

\noindent\textbf{Effect of Different Service Backbones.}
We next evaluate \method\ across different LLM backbones.
As shown in Table~\ref{tab:performance_models} and Figure~\ref{fig:completion_time_models_datasets}, we compare \method\ with the baselines on three different LLMs using OpenOrca as the update source.
Across all tested backbones, \method\ consistently achieves competitive or superior service quality and rollout efficiency.
For example, on Mistral-7B, \method\ achieves a service quality of 88.1\%, compared with 87.4\% for LoRA-Scratch, 87.6\% for PortLLM+FT, and 87.1\% for ORAL.
The efficiency improvement is also substantial.
On LLaMA2-7B, \method\ provides 5.0$\times$ and 2.2$\times$ speedups over LoRA-Scratch and PortLLM+FT, respectively.
These results demonstrate that \method\ is not tied to a specific backbone and can generalize across different backbones.

\noindent\textbf{Hyperparameter Sensitivity Analysis.}
The scheduled regularization strategy introduces two key hyperparameters: $\lambda_1$ for the Guided Rapid Adaptation stage and $\lambda_2$ for the Refinement and Exploration stage.
To evaluate the robustness of \method\ and verify that it does not require extensive per-task tuning, we vary $\lambda_1 \in \{5\text{e}^{-4}, 1\text{e}^{-3}, 5\text{e}^{-3}, 1\text{e}^{-2}\}$ and $\lambda_2 \in \{1\text{e}^{-4}, 5\text{e}^{-4}, 1\text{e}^{-5}\}$ on 20News and MMLU.
The results are summarized in Table~\ref{tab:sensitivity}.
We observe that the final service quality remains stable under different regularization strengths.
For example, on 20News, the service quality stays within a narrow range from 76.48\% to 76.65\%.
However, the choice of $(\lambda_1,\lambda_2)$ affects convergence efficiency.
The configuration $\lambda_1=1\text{e}^{-3}$ and $\lambda_2=1\text{e}^{-4}$ consistently achieves a trade-off between service quality and convergence speed, reaching the target quality in 275 steps on 20News and 625 steps on MMLU.
Therefore, we use this configuration as the default setting in all experiments.
These results suggest that \method\ is robust to hyperparameter variations and avoids the operational burden of sensitive per-service tuning.

\begin{table}[t]
\centering
\caption{Sensitivity of scheduled regularization for service readiness.}
\label{tab:sensitivity}
\vspace{1mm}
\resizebox{\columnwidth}{!}{
\begin{tabular}{cc|cc|cc}
\hline
\multirow{2}{*}{$\lambda_1$} 
& \multirow{2}{*}{$\lambda_2$} 
& \multicolumn{2}{c|}{20News} 
& \multicolumn{2}{c}{MMLU} \\
& 
& Service Quality (\%) 
& Steps to Target Quality 
& Service Quality (\%) 
& Steps to Target Quality \\
\hline
0 & 0 & 76.48 & 356 & 83.11 & 847 \\
$5\text{e}^{-4}$ & $1\text{e}^{-4}$ & 76.61 & 325 & 83.88 & 780 \\
$1\text{e}^{-3}$ & $1\text{e}^{-5}$ & 76.64 & 290 & 83.90 & 711 \\
\textbf{$1\text{e}^{-3}$} & \textbf{$1\text{e}^{-4}$} & 76.63 & \textbf{275} & \textbf{83.92} & \textbf{625} \\
$1\text{e}^{-3}$ & $5\text{e}^{-4}$ & 76.55 & 336 & 83.85 & 642 \\
$5\text{e}^{-3}$ & $1\text{e}^{-4}$ & \textbf{76.65} & 307 & 83.71 & 853 \\
$1\text{e}^{-2}$ & $1\text{e}^{-4}$ & 76.45 & 302 & 83.50 & 827 \\
\hline
\end{tabular}}
\end{table}

\noindent\textbf{Preliminary Extension to Advanced PEFT Variants.}
To examine whether \method\ can extend beyond vanilla LoRA, we apply it to DoRA~\cite{liu2024dora}, a representative LoRA variant that decomposes weight updates into magnitude and direction components.
Following the default setting, we use LLaMA3.1-8B evolved on OpenOrca and evaluate on SST-2.
The results are shown in Table~\ref{tab:dora_res}.
\method\ achieves a service quality of 97.1\%, outperforming DoRA-Scratch and PortLLM+FT, which achieve 96.7\% and 96.9\%, respectively.
In terms of efficiency, \method\ reduces the end-to-end service rollout time to 207 minutes, compared with 517 minutes for DoRA-Scratch and 242 minutes for PortLLM+FT.
These results suggest that the knowledge-reusing design of \method\ can potentially extend to other PEFT variants.

\begin{table}[t]
\centering
\caption{Preliminary extension to DoRA adapters on SST-2.}
\label{tab:dora_res}
\vspace{1mm}
\resizebox{0.85\columnwidth}{!}{
\begin{tabular}{l|c|c}
\hline
Method & Service Quality (\%) & End-to-End Rollout Time (min) \\
\hline
DoRA-Scratch     & 96.7 & 517 \\
PortLLM+FT       & 96.9 & 242 \\
\method\ (Ours)  & \textbf{97.1} & \textbf{207} \\
\hline
\end{tabular}}
\end{table}

\vspace{-3mm}
\section{Related Works}\label{sec:relatedwork}

\subsection{LLM Services and Model Service Maintenance}

LLM-as-a-Service has become an important paradigm for deploying large language models as shared service backbones, where downstream users access model capabilities through APIs, hosted instances, or task-specific adapters. 
In such service ecosystems, base models are periodically updated to incorporate new knowledge, improve safety, or enhance general capabilities, while downstream model services built on previous versions must be maintained accordingly. 
This creates a service lifecycle challenge: updating the base model may invalidate existing task adapters, and retraining numerous adapters from scratch can introduce substantial computation cost and rollout latency. 
Therefore, efficient service maintenance requires mechanisms that can rapidly restore downstream service quality after base-model evolution without incurring full retraining overhead.

\subsection{Parameter-Efficient Fine-Tuning for LLM Services}

Parameter-Efficient Fine-Tuning (PEFT) techniques~\cite{lialin2023scaling,hu2021lora,houlsby2019parameter} have become key enablers for lightweight customization of LLM services. 
Instead of updating all parameters of a large backbone, PEFT freezes the pretrained model and introduces only a small number of trainable parameters, thereby reducing adaptation cost, memory usage, and storage overhead. 
Among existing PEFT methods, LoRA~\cite{hu2021lora} is widely adopted because it injects trainable low-rank matrices into selected layers of the frozen backbone, allowing multiple task-specific adapters to share the same base model. 
This property makes LoRA particularly suitable for multi-tenant LLM service environments, where many downstream services can be deployed as lightweight adapters rather than independent full models. 
However, when the shared backbone evolves, these deployed adapters may no longer remain fully compatible with the updated model, which motivates efficient adapter re-adaptation.

\subsection{Adapting LoRA for Evolving LLMs}

As base LLMs continuously evolve, maintaining the compatibility of previously deployed adapters with updated backbones becomes a critical challenge. 
Retraining each adapter from scratch can recover task performance, but it is expensive and delays service rollout when many downstream services are maintained in production. 
PortLLM~\cite{khan2024portllm} addresses this issue by directly reusing original LoRA weights as transferable ``patches'' for evolved models, which reduces computation cost but may fail to capture the interaction between base-model evolution and task-specific adaptation. 
Another line of work focuses on generating LoRA parameters from task or model descriptions. 
For example, Cond P-Diff~\cite{jin2024conditional} and ORAL~\cite{khan2025oral} use conditional diffusion-based models to synthesize LoRA weights for target tasks and architectures. 
Although these approaches provide flexibility in creating adapters on demand, they typically require substantial task-adapter training data, and their effectiveness depends on the quality of the learned generator and conditioning signals. 
In contrast, \method\ focuses on re-adapting existing deployed adapters by explicitly reusing both the old task-specific adapter and the parameter delta introduced by base-model evolution, enabling efficient service-quality recovery after model updates.

\subsection{Efficient Service Rollout and Model Update}

Efficient model update and service rollout are also related to model merging and version-aware deployment. 
Model merging aims to combine useful capabilities from different models or adapters without full retraining~\cite{yang2024model}. 
Representative methods include arithmetic strategies such as Task Arithmetic~\cite{ilharco2022editing}, statistical approaches such as Fisher-Merging~\cite{matena2022merging} and RegMean~\cite{jin2022dataless}, and search-based methods such as AdaMerging~\cite{yang2023adamerging} and Evolutionary Model Merge~\cite{akiba2025evolutionary}. 
These methods are mainly designed to merge task vectors or adapters for multi-task learning on a relatively static backbone, where the goal is to balance capabilities from multiple tasks. 
By contrast, \method\ addresses a different service maintenance problem: the base model changes over time, and the objective is to realign an existing task adapter with the evolved backbone. 
Rather than merging multiple task adapters, \method\ uses Bayesian optimization to identify task-specific fusion coefficients between the old adapter and the base-evolution delta, thereby reducing rollout latency while preserving service quality.

\vspace{-3mm}
\section{Discussion and Limitation}\label{sec:limitation}

\subsection{Implications for LLM Service Ecosystems}
Analogous to software engineering where operating system updates ideally maintain compatibility with existing applications, the evolution of base LLMs poses a significant backward compatibility challenge. Currently, downstream consumers bear the full cost of retraining adapters for every upstream update. ReLoRA addresses this by serving as a bridge between the parameter spaces of the old and new models. We envision a future ecosystem where providers release compatibility metadata alongside updates, allowing ReLoRA to transform adaptation from an expensive retraining task into a near-instantaneous patching process.

\subsection{Broader Application Scenarios}
While our current evaluation focuses on continual pre-training, practical LLM evolution encompasses diverse stages, such as Supervised Fine-Tuning (SFT) and Reinforcement Learning from Human Feedback (RLHF). 
Although these alignment processes induce parameter shifts that are distinct from knowledge injection, the core principle of ReLoRA remains applicable.
By conceptualizing the parameter difference as a generic evolution vector, our framework can readily adapt to these scenarios or even support quantization-aware updates. 
Consequently, ReLoRA holds the potential to serve as a unified protocol for preserving task-specific capabilities across the entire model lifecycle.

\subsection{Limitations of Offline Evaluation}
\textbf{Online serving metrics.} Our evaluation focuses on offline adapter re-adaptation time and service-quality recovery. It does not directly measure online SLA violations, request-level serving latency, or multi-tenant serving interference during deployment. These system-level serving metrics are orthogonal to ReLoRA's adapter update mechanism and will be explored in future work.



\subsection{Technical Limitations and Future Work}
While ReLoRA achieves significant improvements in adapting LoRA adapters for evolving LLMs, we acknowledge several limitations that motivate our future research. 
1) \textbf{Coarse-grained fusion strategy.} 
Our core fusion mechanism, $\alpha \cdot \Delta \Theta +\beta \cdot \Delta \Theta_i$, currently employs global scalar coefficients $\alpha$ and $\beta$. 
This design relies on a simplified assumption of uniform interplay between model evolution and task adaptation across all layers.
However, this overlooks the hierarchical nature of LLMs, where different layers or module types (\eg, Attention vs. FFNs) may benefit from distinct fusion strategies.
For instance, lower layers processing syntax may require different fusion ratios compared to upper layers handling semantics. 
Future work could explore more fine-grained methods, such as layer-wise fusion strategies.
2) \textbf{Reliability of the proxy objective.} 
\method currently utilizes the initial validation loss as a proxy to guide the Bayesian search.
While computationally efficient, the initial loss is not always a reliable predictor of the final model  performance. 
 For instance, a configuration with slightly higher initial loss might reside in a wider and flatter basin, potentially leading to superior generalization after fine-tuning. 
To address this limitation, we plan to incorporate geometric metrics, such as Hessian-based approximations, to identify more robust initialization candidates in future work.

\setlength{\parindent}{10pt}

\section{Conclusion}\label{sec:concluesion}
In this paper, we propose \method, a service maintenance framework for fast rollout of evolving LLM services. 
We formulate adapter re-adaptation as a service rollout and adapter backward-compatibility problem, where deployed task-specific adapters must be efficiently updated after base-model evolution. 
To this end, \method\ reuses two forms of knowledge: the task-specific knowledge in the deployed adapter and the evolution knowledge captured by the base-model parameter delta. 
Specifically, \method\ constructs a compatibility-aware starting point through adaptive LoRA initialization, and then applies scheduled regularization to accelerate service-readiness recovery while preserving final task quality. 
Extensive experiments across six downstream service tasks, three model families, and three update sources demonstrate that \method\ reduces time-to-readiness by up to 8.9$\times$ and improves service quality by up to 4.6 percentage points over baselines. 
These results show that \method\ effectively reduces adapter maintenance overhead and accelerates the rollout of updated LLM services.

\vspace{-4mm}
\bibliographystyle{IEEEtran}
\bibliography{refs}

\end{document}